# TianHui: A Domain-Specific Large Language Model for Diverse Traditional Chinese Medicine Scenarios

Ji Yin[1, 2, +]; Menglan He[1, +]; Yujie Zhang[1]; Linshuai Zhang[1]; Tingting Ma[3]; Ce Tian[3]; Jie Wu[3]; Lin Xu[1, *]; Tao Jiang[1, 2, *]

[1]. School of Intelligent Medicine, Chengdu University of Traditional Chinese Medicine, Chengdu, China.

[2]. The Acupuncture and Tuina School, Chengdu University of Traditional Chinese Medicine, Chengdu, China.

[3]. Center of Preventive Medicine, Hospital of Chengdu University of Traditional Chinese Medicine, Chengdu, China.

[+] These authors have contributed equally to this work.

[*] Corresponding author:

Tao Jiang, MD

School of Intelligent Medicine

Chengdu University of Traditional Chinese Medicine

No.1166, Liutai Avenue

Wenjiang District

Chengdu, 610000

China

Phone: 86 13308081686

Email: jiangtop@cdutcm.edu.cn

Lin Xu, MD

School of Intelligent Medicine

Chengdu University of Traditional Chinese Medicine

No.1166, Liutai Avenue

Wenjiang District

Chengdu, 610000

China

Phone: 86 18780206202

Email: xulin@cdutcm.edu.cn

**Abstract**

**Background:** Currently, domain-specific large language models (LLMs) in traditional Chinese medicine (TCM) are primarily designed for clinical practice and medical education, yet they demonstrate substantial limitations when applied to research contexts owing to inadequate adaptability to complex tasks, thereby constraining their scientific utility. Moreover, the absence of comprehensive evaluation datasets and computational resource constraints hinder rigorous performance assessments and prevent extensive comparative or ablation experiments, ultimately resulting in suboptimal model performance and weakened persuasiveness.

**Objective:** To address these challenges, this study proposed a method for constructing a specialized LLM for the TCM domain based on contextual data integration and domain knowledge fusion and successfully developed a privatized LLM for the TCM profession, TianHui.

**Methods:** Firstly, we acquired a large amount of TCM data, including academic literature resources, published book materials, online public data, and other supplementary materials, and pre-processed them to finally generate the 0.97G unsupervised dataset and 611312 QAs. Then, we adopted a phased training strategy (Pre-Training (PT) and Supervised Fine-Tuning (SFT)) and integrated three key technologies, Quantized Low-Rank Adaptation (QLoRA) parameter efficient fine-tuning, DeepSpeed Stage 2 distributed training optimization, and Flash Attention 2 accelerated computation, to achieve optimal allocation of computational resources while guaranteeing training stability. Finally, we evaluated TianHui using 12 different types of benchmark test datasets and conducted extensive comparison experiments and ablation experiments.

**Results:** The benchmark test data showed that TianHui demonstrated excellent performance in 12 TCM-related application scenarios. It ranked in the top three in each evaluation index in six test datasets: APQ, TCMCD, HFR, HCCA, DHPE, and TLAW. Meanwhile, it achieved optimal performance in all indicators of the six test data sets: TCMEE, APR, GCPMI, TCMKQA, TCMRC, and ADTG. To evaluate the effect of each hyper-parameter on the performance of TianHui, we gradually ablated LoRA rank (8, 16, 32, 64, 128), LoRA alpha (16, 32, 64, 128, 256), dropout (0, 0.2, 0.4), epoch (2, 4, 6), and max length (256, 512, 1024, 2048). Finally, the hyper-parameters of TianHui were determined to be LoRA rank of 128, LoRA alpha of 256, epoch of 4, dropout of 0.2, and max length of 2048.

**Conclusions:** In conclusion, we successfully developed TianHui, a professional, privatized LLM for the TCM domain. Through continuous technology iteration and accumulation of TCM featured data, TianHui not only significantly improved the accuracy and professionalism of the TCM knowledge process, but also provided an intelligent solution for the systematic inheritance and large-scale application of TCM knowledge. Our code, data, and models are all open-sourced on GitHub (https://github.com/JYfantast/TianHui) and HuggingFace (https://huggingface.co/JYfantast/TianHui).

## Introduction

In recent years, the research on fine-tuning pre-trained language models (PLMs) received extensive attention in the field of natural language processing (NLP) [1-2]. After training on a large amount of textual data, PLMs demonstrated a remarkable ability to capture deep semantic insights, significantly enhancing the performance of numerous NLP tasks [1-2]. Initially, BERT employed a transformer encoder with multi-head attention mechanisms, pre-trained on large-scale corpora, which substantially improved downstream task performance [4]. More recently, large language models (LLMs), exemplified by ChatGPT, garnered increasing attention. Building upon GPT-3, these LLMs incorporated advanced techniques, including instruction tuning and reinforcement learning from human feedback (RLHF), exhibiting exceptional text-generation capabilities [5]. General-purpose LLMs such as Qwen2, BaiChuan2, and DeepSeek demonstrated impressive performance across a wide range of tasks [6-7]. These LLMs were able to interpret human intentions and produce contextually appropriate responses, handling tasks ranging from simple text generation to advanced applications like dialogue systems and domain-specific reasoning (e.g., in healthcare, law, and finance)[6-7]. Among these fields, healthcare emerged as a key research focus due to its critical impact on human health and longevity, with traditional Chinese medicine (TCM), in particular, representing a challenging yet highly promising research direction.

Several notable LLMs were developed with specialized applications in the field of TCM, demonstrating significant advancements in medical artificial intelligence (AI). CMLM-ZhongJing underwent multiple iterative training cycles on proprietary medical datasets, employing structured professional templates and tailored prompt designs through a comprehensive training pipeline that encompassed pre-training (PT), supervised fine-tuning (SFT), and RLHF to acquire advanced reasoning capabilities for TCM prescription formulation and diagnostic logic [11]. BianQue utilized ChatGLM-6B as its foundational architecture, integrating a Chinese medical Question and Answer (Q&A) dataset containing tens of millions of entries with proprietary health dialogue data through multiple rounds of inquiry-based and instructional fine-tuning, while employing full fine-tuning (FFT) to effectively simulate physician consultation processes［12］. Lingdan was developed based on the Baichuan2-13B-Base, employing PT on large-scale TCM-domain corpora including classical texts, modern textbooks, and the Chinese pharmacopoeia, followed by supervised fine-tuning using annotated TCM-specific datasets［13］. Huatuo incorporated three distinct data sources, medical instruction data distilled from ChatGPT, authentic physician instruction records, and Q&A pairs structured through medical knowledge graphs, and employed a two-phase training process that initially utilized SFT with 40,000 verified medical question-answer pairs to establish medical reasoning pathways, followed by reinforcement learning optimization using verifier-based reward mechanisms[14]. TCMChat was developed based on the Baichuan2-7B-Chat architecture and, through combined PT and SFT phases, incorporated large-scale TCM textual knowledge with Chinese Q&A datasets for comprehensive training[15]. LLMs for TCM transformed medical AI by integrating advanced training methods with extensive TCM knowledge, enabling advances in prescription generation, diagnostics, and consultation simulation.

In the domain of TCM, while several LLMs have been developed and demonstrated measurable efficacy in both clinical practice and medical education, significant limitations and challenges remain evident in their current implementations. Firstly, current TCM LLMs primarily focused on clinical practice and medical education, while exhibiting substantial limitations in research-oriented contexts. The developed LLMs demonstrated insufficient adaptability when applied to complex research tasks, which consequently restricted their utility in scientific inquiry and constrained their potential research applications. Secondly, the absence of a comprehensive evaluation dataset compromised the persuasiveness of existing assessment outcomes, potentially leading to overestimation or underestimation of LLMs' performance in TCM-related tasks. Finally, constrained by computational resource limitations, numerous studies failed to conduct extensive comparative and ablation experiments, ultimately resulting in poor model performance as well as poor persuasion.

To address these challenges, this study proposed a method for constructing a specialized LLM for the TCM domain based on contextual data integration and domain knowledge fusion and successfully developed a privatized LLM for TCM profession, TianHui. Firstly, we acquired a large amount of TCM data, including academic literature resources, published book materials, online public data and other supplementary materials, and pre-processed them to finally generate the 0.97G unsupervised dataset and 611312 QAs. Then, we adopted a phased training strategy (PT and SFT) and integrated three key technologies, Quantized Low-Rank Adaptation (QLoRA) parameter efficient fine-tuning, DeepSpeed Stage 2 distributed training optimization and Flash Attention 2 accelerated computation, to achieve optimal allocation of computational resources while guaranteeing training stability. Finally, we evaluated TianHui using 12 different types of benchmark test datasets and conducted extensive comparison experiments and ablation experiments. Through continuous technology iteration and accumulation of TCM featured data, TianHui not only significantly improved the accuracy and professionalism of TCM knowledge process, but also provided an intelligent solution for the systematic inheritance and large-scale application of TCM knowledge.

## Methods and datasets

Collection of data

The study adopted a multi-source heterogeneous data integration approach to obtain high-quality research data through a systematic data collection and processing process. Specifically, our data sources mainly included the following four categories: academic literature resources, published book materials, online public data and other supplementary materials. The study systematically searched two core academic resource platforms, PubMed (\url{https://pubmed.ncbi.nlm.nih.gov}), an international authoritative biomedical database, and China National Knowledge Infrastructure (CNKI, \url{https://www.cnki.net}), and adopted a professional search strategy to obtain research literature related to TCM. By setting strict inclusion and exclusion criteria, the information (including titles and abstracts) of 142,178 high-quality TCM research literatures was finally screened and obtained. The corpus of this study was systematically compiled from authoritative medical sources, including nationally recognized standards, peer-reviewed medical textbooks, and clinically documented medical cases. Specifically, the dataset comprises: 41 medical textbooks and 1522 TCM ancient books covering foundational theories, diagnostic methods, and therapeutic principles in traditional and modern medicine; 18 rigorously documented medical cases representing diverse clinical scenarios and treatment outcomes; 4 national standards (e.g., pharmacopoeias) to ensure compliance with regulatory and evidence-based medical frameworks (\url{https://24hbook.daohangxie.com} and \url{https://github.com/ZJUFanLab/TCMChat}) [13, 15]. ChP-TCM was a TCM dataset containing 13,006 TCM-related data records [13, 15]. ShenNong-TCM was a large-scale professional dataset of online Chinese medicine consultations, featuring 112,565 real-world Chinese medical consultation dialogue records [17]. Chinese Patent Medicine Instructions (CPMI) 3906 cases, mainly from the National Standard Treatment Guidelines for Essential Medicines, TCM instruction records from the Tianchi dataset and real-world Chinese Patent Medicine (CPM) prescriptions from 100 patients[18]. Lingdan dataset was a TCM dataset integrating 150,675 TCM basic knowledge and clinical diagnosis and treatment data . TCM-SD is a professional dataset of Chinese medicine discernment containing 49,693 structured discernment records [19]. The TCM Knowledge Base is a TCM database that integrates 606,500 structured data to support multi-task learning and research in the TCM domain [19]. A total of 1,406 TCM basic knowledge data were collected by collating unstructured text data from authoritative TCM books.

## Collation of data

Unsupervised data were primarily obtained from books, CNKI, pubmed, TCM-SD, TCMChat, ChP-TCM, ShenNong-TCM and LingdanLLM. Firstly, we obtained the books in EPUB or PDF format, for EPUB format, the text information was extracted directly, for PDF format, the text information was extracted using OCR. All these text information required manual proofreading and editing. Secondly, for the academic literature data obtained from CNKI or pubmed, we deleted the failed crawling, empty and duplicate data, and then after manual checking, we only retained the data related to TCM. Finally, for the data from public datasets, we deleted the parts containing modern medicine, removed duplicates, and standardized the format. The supervision instruction data was mainly sourced from CPMI-ChatGLM, TCMChat and books and its construction methodology was a systematic process to ensure that TianHui was able to perform specific tasks. Specifically, we used different construction strategies: 1) TCM Knowledge Linguisticisation; 2) Extract Structured Data; and 3) Instruction Data Refinement. The data from these strategies are filtered through manual verification to generate 12 basic scenarios. The evaluation datasets were obtained from TCMChat, CPMI-ChatGLM,

CNKI, pubmed and books. Based on clinical, instructional, and research scenarios, 12 usage scenarios were finally constructed, including answer prediction question (APQ), TCM case diagnosis (TCMCD), TCM entity extraction (TCMEE), herb or formula recommendation (HFR), acupuncture point recommendation (APR), herbal chemical composition analysis (HCCA), generation of Chinese patent medicine instruction (GCPMI), description of herbal pharmacological Effect (DHPE), TCM knowledge questions and answer (TCMKQA), TCM reading comprehension (TCMRC), topic-led abstract writing (TLAW), and abstract-driven topic generation (ADTG). All data underwent a strict quality control process, including de-duplication and cleaning, format standardization, and other processing steps, to ensure the reliability and completeness of the data.

## Presentation of the base model

The TianHui LLM was developed based on DeepSeek-R1-Distill-Qwen-14B, a distilled model that was optimized from the Qwen2.5-14B architecture through dynamic sparse attention and layer-wise knowledge distillation techniques to enhance domain-specific semantic understanding while maintaining computational efficiency [19], and [21]. We initially tried the DeepSeek-R1-Distill-Qwen-7B version of the model, but due to its shortcomings in basic linguistic expressiveness and generalizability to complex tasks, we finally chose the DeepSeek-R1-Distill-Qwen-14B version as the base model.

## Model Training

During the model training phase, we implemented a two-stage process consisting of PT and SFT within the LLaMA-Factory architectural framework[22]. For the PT phase, the model performs unsupervised learning on large-scale textual data to capture deep semantic information. Specifically, we used DeepSpeed Stage 2 to achieve efficient distributed training and combined Flash Attention 2 technology to accelerate the computation of attention mechanisms, significantly improving training efficiency and model convergence speed[22]. In addition, to address the memory bottleneck problem in large-scale model training, we introduced QLoRA technology, which significantly reduced memory usage and computational requirements by quantifying model weights, enabling efficient PT even with limited hardware resources [25].

In the SFT stage, we used a high-quality annotated dataset for supervised training of the model to further improve its performance on specific tasks. Similar to the PT stage, we also adopted DeepSpeed Stage 2 and Flash Attention 2 techniques in the SFT stage to ensure the efficiency and stability of the training process. Specifically, to further enhance the adaptability and generalization ability of the model, we used instruction data to enhance the adaptability of the deep learning model. Through the comprehensive application of these technologies, we can not only ensure model performance while significantly reducing training time, but also effectively reduce the consumption of computing resources, thereby achieving efficient training of large-scale deep learning models.

## Performance evaluation

To compare TianHui's performance with other LLMs in TCM application scenarios, 12 domain-specific benchmark test datasets were developed, which covered various TCM scenarios such as APQ, TCMCD, TCMEE, HFR, APR, HCCA, GCPMI, DHPE, TCMKQA, TCMRC, TLAW, and ADTG. Model performance was evaluated using more comprehensive metrics.

### Evaluation on answer prediction questions

2000 single choice questions were used as APQ test dataset and accuracy was calculated using the following formula:

$$\text{Accuracy} = \frac{TP + TN}{TP + TN + FP + FN}$$

Where:

- TP (True Positives): The number of correctly predicted positive instances.
- TN (True Negatives): The number of correctly predicted negative instances.
- FP (False Positives): The number of incorrectly predicted positive instances.
- FN (False Negatives): The number of incorrectly predicted negative instances.

### Evaluation on TCM case diagnosis

The TCMCD test set contained 500 data and evaluated them using accuracy.

### Evaluation on TCM entity extraction

The TCMEE test set comprised 480 data points, and precision, recall, and F1 scores were employed to assess them. The F1-score, which balanced precision (the accuracy of positive predictions) and recall (the identification of true positive instances), was utilized to gauge model performance. It encompassed two fundamental metrics: precision and recall. The former quantified the matching degree between the words in the generated text and those in the reference text, whereas the latter assessed the matching degree between the words in the reference text and those in the generated text. The formula was as follows:

$$\text{Precision} = \frac{TP}{TP + FP}$$
$$\text{Recall} = \frac{TP}{TP + FN}$$
$$F1 = 2 \times \frac{\text{Precision} \times \text{Recall}}{\text{Precision} + \text{Recall}}$$

Where:

- TP (True Positives): The number of correctly predicted positive instances.
- FP (False Positives): The number of incorrectly predicted positive instances.
- FN (False Negatives): The number of incorrectly predicted negative instances.

### Evaluation on herb or formula recommendation

The HFR test set consisted of 500 data and was evaluated using the Mean Reciprocal Rank (MRR)[25], precision@3, recall@3, HR@3 and Normalized Discounted Cumulative Gain (nDCG)[27].

The MRR served as a metric for evaluating the performance of an information retrieval system. It was employed to appraise the ranking quality of relevant documents within the system's output. The formula is as follows:

$$\mathrm{MRR} = \frac{1}{|Q|} \sum_{i=1}^{|Q|} \frac{1}{\mathrm{rank}_i}$$

where |Q| is the number of recommendations and rank i is the i-th recommendation. The Hit Rate (HR) measures whether the recommended items contain the user's actual interactions. It evaluates the system's ability to include relevant items in the recommendation list, focusing on prediction accuracy. The formula is as follows:

$$\mathrm{HR} = \frac{1}{N} \sum_{i=1}^{N} \mathbb{I}(\mathrm{item}_i \in \mathrm{recommended\ list}),$$

where N is the number of users (or test cases), and I (·) is an indicator function that equals 1 if the user's interacted item exists in the recommendation list, otherwise 0. The precision@k indicated the system's capability to offer correct or relevant information within the first K referral or search results. The formula is as follows:

$$\mathrm{Precision@}K = \frac{\mathrm{TP@}K}{\mathrm{TP@}K + \mathrm{FN@}K}$$

where K denotes taking the first K recommendations.

The nDCG was a metric that evaluated the quality of search results and was employed to assess the quality of search results or recommendation lists. It took into account the relevance of each item and its position (rank) in the search results, with the position being discounted due to the fact that users tended to focus more on the top-ranked results. The formula is as follows:

$$\mathrm{nDCG} = \frac{\sum_{i=1}^{p} \frac{2^{\mathrm{rel}_i} - 1}{\log_2 (i + 1)}}{\sum_{i=1}^{|REL|} \frac{2^{\mathrm{rel}_i} - 1}{\log_2 (i + 1)}}$$

Where reli represented the relevance of the recommendation result at position i, p denoted the size of the recommendation list, and |REL| indicated the set of all sorted recommendations.

### Evaluation on acupuncture point recommendations

The APR test set consisted of 350 data and was evaluated using the accuracy, nDCG, precision@K, recall@K, and HR@K.

### Evaluation on TCM knowledge questions and answers

The TCMKQA test set was composed of 500 data points and assessed utilizing the Bilingual Evaluation Understudy (BLEU)[28], Metric for Evaluation of Translation with Explicit Ordering (METEOR)[28], Recall-Oriented Understudy for Gisting Evaluation (ROUGE)[30], and BertScore.

The BLEU metric was utilized to measure the similarity between generated and reference sentences through the assessment of n-gram overlap, which served to evaluate sentence-level fluency. The formula is as follows:

$$\text{BLEU} = BP \times \exp\left(\sum_{n=1}^{N} W_n \times \log P_n\right)$$

Where:

- $W_n$ indicates the weights assigned to the n-gram,

- BP represents the penalty factor,

- $P_n$ denotes the precision of the n-gram.

The METEOR was employed as a software metric to evaluate text quality and complexity. The formula is as follows:

$$F = \frac{P \times R}{\alpha \times P + (1-\alpha) \times R}$$

$$\text{Pen} = \gamma \times \left(\frac{m}{ch}\right)^{\beta}$$

$$\text{METEOR} = (1 - \text{Pen}) \times F$$

Where:

- P and R respectively correspond to the precision and recall of the generated text,

- m signifies the quantity of unbroken sequences of consecutive matching words between the generated text and reference text,

- ch indicates the count of unigrams that match in the generated text and reference text,

- γ and β serve as hyper-parameters.

The ROUGE metric was employed to assess the overlap of n-grams between generated outputs and reference summaries. It utilized ROUGE-1, ROUGE-2, and ROUGE-L to evaluate the quality of model responses based on word matches in the longest common sub-sequence. The formula is as follows:

$$\text{ROUGE} - \text{N} = \frac{\sum_{S \in \text{ReferenceSummaries}} \sum_{\text{gram}_n \in S} \text{Count}_{\text{match}}(\text{gram}_n)}{\sum_{S \in \text{ReferenceSummaries}} \sum_{\text{gram}_n \in S} \text{Count}(\text{gram}_n)}$$

Where:

• Reference summaries refers to the reference text,

• Count match (gramn) indicates the number of n-grams that appear in both the generated text and the reference text,

• Count (gramn) denotes the number of n-grams appearing in the reference text.

ROUGE-L is calculated as:

$$\text{ROUGE} - \text{L} = \frac{(1 + \beta^2) \times \text{LCS}(x, \hat{x})}{\text{len}(x) + \beta^2 \times \text{len}(\hat{x})}$$

Where:

• x represents the reference text,

• x̂ corresponds to the generated text,

• LCS(x, x̂) denotes the length of the longest common subsequence between x and x̂.

*Evaluation on generation of Chinese patent medicine instructions*

The GCPMI test set consisted of 566 data and was evaluated using the BLEU, METEOR, ROUGE and BertScore.

*Evaluation on TCM reading comprehension*

The 500 data were used as TCMRC test set and their similarity was assessed using BLEU, METEOR, ROUGE and BertScore.

*Evaluation on description of herbal pharmacological effects*

The DHPE test set consisted of 437 data and was evaluated using the BLEU, METEOR, ROUGE and BertScore.

*Evaluation on herbal chemical composition analysis*

The HCCA test set consisted of 437 data and was evaluated using the BLEU, METEOR, ROUGE and BertScore.

*Evaluation on abstract-driven topic generation*

The ADTG test set consisted of 1000 data and was evaluated using the BLEU, METEOR, ROUGE and BertScore.

*Evaluation on topic-led abstract writing*

The TLAW test set consisted of 1000 data and was evaluated using the BLEU, METEOR, ROUGE and BertScore.

**Comparison with other LLMs**

To comprehensively evaluate the performance of TianHui against existing LLMs, we selected a series of LLMs with different parameters as comparison baselines, including general, Chinese medical, and TCM-specific LLMs. The specific comparison models were shown in Table 1.

Table 1: Information about the 27 LLMs, including model name, base model, model type (TCM, Chinese medical, and general) and deployment method (local and API).

| Model | Base model | Model type | Deployment method |
| --- | --- | --- | --- |
| Bentsao | LLaMA-7B | TCM | Local |
| BianCang-Qwen2.5-7B-Instruct | Qwen2.5-7B-Instruct | TCM | Local |
| HuatuoGPT2-13B | Baichuan2-13B-Base | TCM | Local |
| Lingdan-13B-Base | Baichuan2-13B-Base | TCM | Local |
| Lingdan-13B-PR | Baichuan2-13B-Base | TCM | Local |
| ShenNong | LlaMA-7B | TCM | Local |
| Sunsimiao-7B | Qwen2-7B | TCM | Local |
| TCMchat | baichuan2-7B-Chat | TCM | Local |
| ZhongjingGPT1-13B | Baichuan2-13B-Chat | TCM | Local |
| Baichuan-M1-14B | Baichuan-M1-14B | Chinese medical | Local |
| BianQue-2 | ChatGLM-6B | Chinese medical | Local |
| Chatmed | Llama-7b | Chinese medical | Local |
| Baichuan2-13B-Chat | Baichuan2-13B-Chat | General | Local |
| Baichuan2-7B-Chat | Baichuan2-7B-Chat | General | Local |
| ChatGLM3-6B | ChatGLM3-6B | General | Local |
| DeepSeek-R1-Distill-Qwen-14B | Qwen2.5-14B | General | Local |

| DeepSeek-R1-Distill-Qwen-32B | Qwen2.5-32B | General | Local |
| --- | --- | --- | --- |
| DeepSeek-R1-Distill-Qwen-7B | Qwen2.5-Math-7B | General | Local |
| Llama3-8B-Chinese-Chat | Llama3-8B-Chinese-Chat | General | Local |
| Qwen2.5-14B-Instruct | Qwen2.5-14B | General | Local |
| Qwen2.5-32B-Instruct | Qwen2.5-32B | General | Local |
| Qwen2.5-72B-Instruct | qwen2.5-72B | General | Local |
| Qwen2.5-Math-7B | Qwen2.5-Math-7B | General | Local |
| Deepseek-R1 | Deepseek-R1 | General | API |
| Deepseek-V3 | Deepseek-V3 | General | API |
| GPT-3.5-turbo | GPT-3.5-turbo | General | API |
| GPT-4o | GPT-4o | General | API |

**Ablation study**

To evaluate the effect of each hyper-parameter on the performance of the model, we gradually ablated LoRA rank (8, 16, 32, 64, 128), LoRA alpha (16, 32, 64, 128, 256), dropout(0, 0.2, 0.4), epoch (2, 4, 6), and max length (256, 512, 1024, 2048). In each round of experiment, only one parameter value was changed, and the other parameters remained the default settings. By comparing the performance evaluation indicators of the model under different settings, the role and impact of each parameter were analyzed.

Specifically, in the LoRA rank ablation experiment, we will set the rank to 8, 16, 32, 64 and 128 respectively, and observe how the performance of the model changes with the increase of the rank value; In the LoRA alpha ablation experiment, we adjusted the alpha values to 16, 32, 64, 128, 256, and analyzed the effects of different alpha values on the expression ability of the model. In the dropout ablation experiment, we set the dropout rate to 0, 0.2 and 0.4, respectively, to study its effect on the generalization ability and over fitting phenomenon of the model. In the epoch ablation experiment, we set the number of training epochs to 2, 4 and 6 respectively, and observed the effect of different training rounds on the performance of the model, so as to determine the optimal training duration. In the maximum length ablation study, we evaluated the impact on model performance by testing values of 256, 512, 1024, and 2048.

### Computational hardware and software

Both PT and SFT experiments were performed on 8 NVIDIA A100 40G GPUs. The specific python version and related modules referred to the LLaMA-Factory project. The deployment of the comparison model and dialogue generation referred to the corresponding official website.

## Results

### Overview of TianHui datasets

In this research, details of the PT and SFT datasets used in TianHui were presented in Table 2. We used the four processing methods shown in Figure 1 for (A) TCM

Knowledge Linguisticisation; (B) Extract Structured Data; (C) Instruction Data Refinement. For the PT datasets, we obtained 203 MB of academic literature resources from CNKI and PubMed, 39 MB of book resources from TCMChat and 24hbook, and 0.94 G of TCM-related text data from ShenNong-TCM, BianCang, Lingdan, TCMChat, and TCM-SD, and finally constructed a roughly 0.97 G unsupervised dataset. For the SFT dataset, we obtained 611312 QAs pairs from TCMChat and CPMI-ChatGLM.

Table 2: Details of the PT and SFT datasets used in TianHui.

| Data type | Content | Size | Source |
|---|---|---|---|
| PT | Academic literature resources | 203 MB | CNKI, PubMed |
| PT | Book | 39 MB | TCMChat, 24hbook |
| PT | TCM-text | 0.9 G | ShenNong-TCM, BianCang, Lingdan, TCMChat, TCM-SD |
| SFT | TCM-QAs | 611312 QAs | TCMChat, CPMI-ChatGLM |

Figure 1 Overview of construction for instruction data.

(A) TCM Knowledge Linguisticisation. We transformed the language expression of TCM knowledge from books into colloquial language using GPT-3.5-turbo. (B) Extract Structured Data. High-quality instruction sets were obtained from open-source databases through automatic and manual filtering to enhance the performance of the models on unseen data. (C) Instruction Data Retrieval. Teaching data from various sources, including APQ, TCMCD, TCMEE, HFR, APR, HCCA, GCPMI, DHPE, TCMKQA, TCMRC, TLAW, and ADTG, were first compiled. Subsequently, the instruction set was converted into a data format recognizable by the LLM. Ultimately, the final training and testing sets were obtained through random sampling.

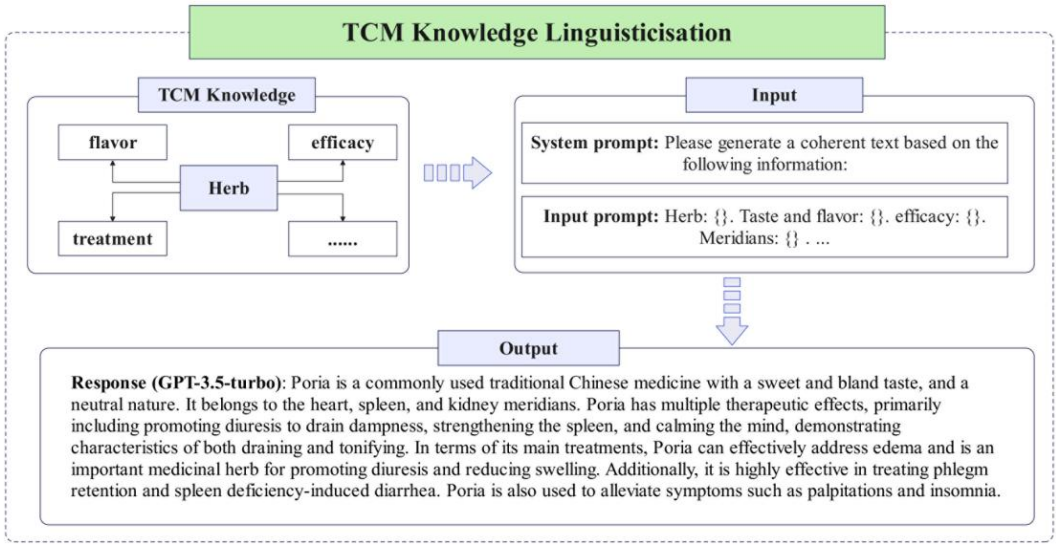
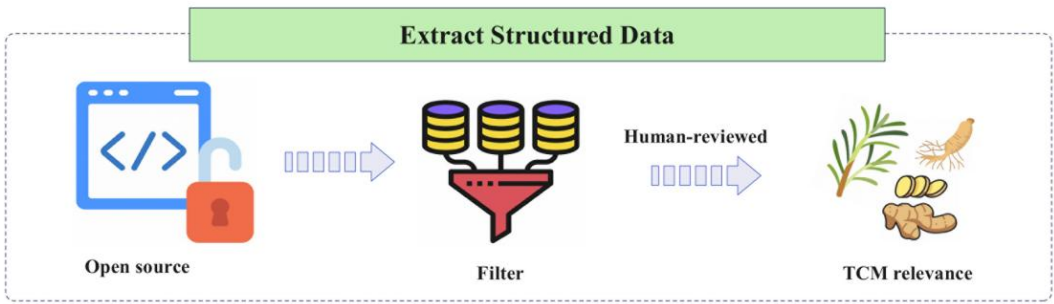
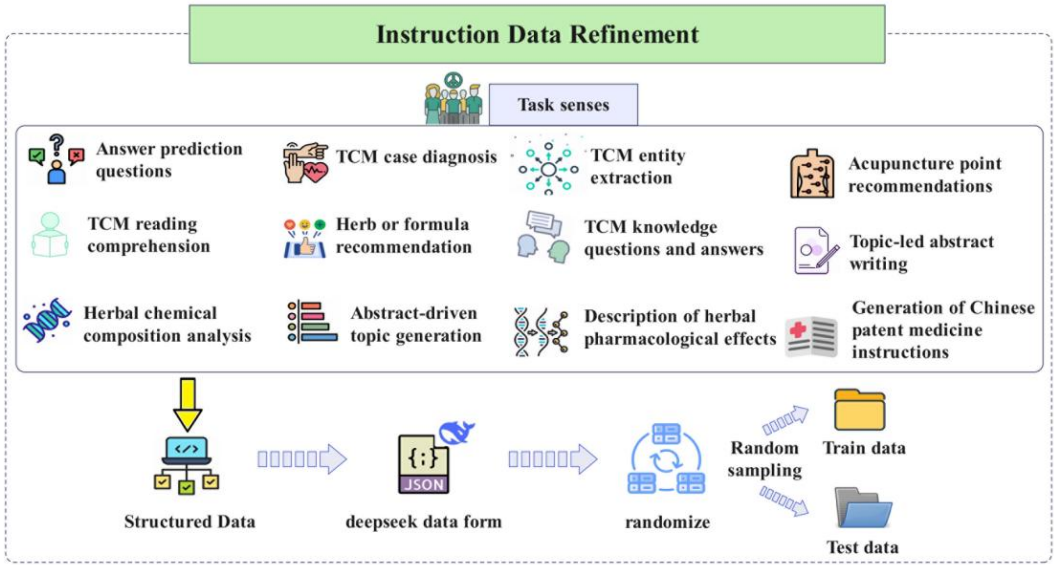

## Workflow of TianHui model

Before training the models, we comprehensively evaluated eight generic models (Baichuan2-7B-Chat, Baichuan2-13B-Chat, ChatGLM3-6B, DeepSeek-R1-Distill-Qwen-7B, DeepSeek-R1-Distill-Qwen-14B, Qwen2.5-Math-7B, Qwen2.5-14B-Instruct, Llama3-8B-Chinese-Chat) to select the most appropriate baseline model. In our experiments, the 13B or 14B parameter models demonstrated significantly better comprehension and overall performance compared to 7B or 8B parameter models, exhibiting superior language understanding, reasoning capabilities, and task performance across various benchmarks, with DeepSeek-R1-Distill-Qwen-14B showing the highest performance among the 13B-14B models (Figure 3-5). Therefore, DeepSeek-R1-Distill-Qwen-14B was identified as the base model for TianHui. Figure 2 illustrated the research process of the study.

Figure 2 The construction process of TianHui included data preparation, model training, and evaluation.

Data preparation primarily entailed the processes of collection and processing. The raw data collected, which mainly originated from books, open-source databases, crawlers, and literature, was constructed into unsupervised data and supervised instruction sets through various strategies. TianHui used the DeepSeek-R1-Distill-Qwen-14B pedestal model, which was obtained through two stages of PT and SFT. The performance of TianHui was evaluated using 12 test datasets from different scenarios. The evaluation indicators mainly included accuracy, precision, recall, F1, MRR, HR, nDCG, BLEU, ROUGE, METEOR, and BERTScore.

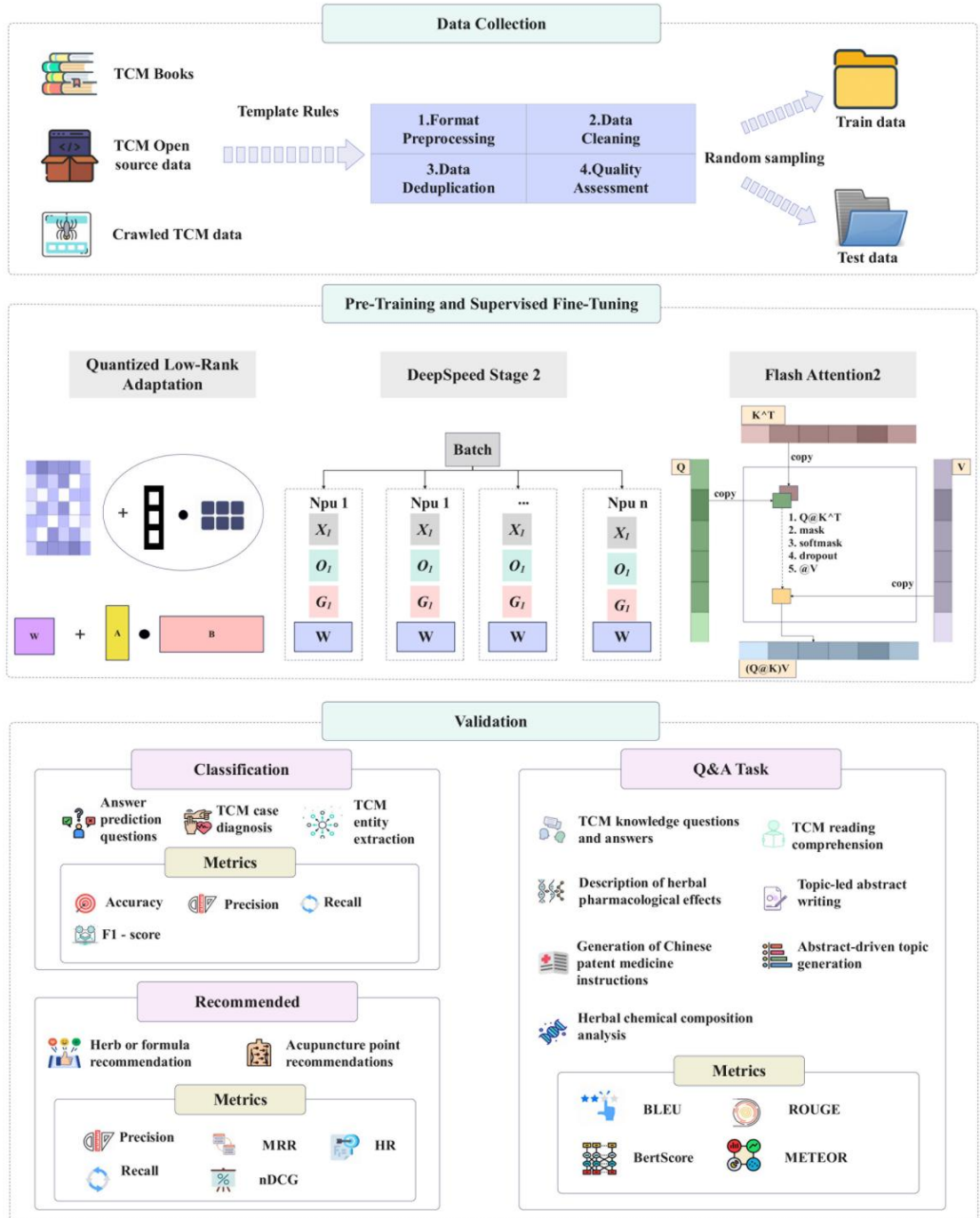

Our model training protocol employed a optimized configuration to achieve an optimal balance between computational efficiency and model performance. Specifically, we implemented QLoRA (4 bit) with a rank of 128 and an alpha parameter of 256, applying a dropout rate of 0.2 to prevent over-fitting. The training regimen consisted of 4.0 complete epochs with a maximum sequence length of 2048 tokens. We configured a per-device batch size of 4 samples with 8 gradient accumulation steps, enabling effective batch processing while maintaining memory efficiency. The optimization process utilized a learning rate of $5.0\times10^{-5}$ with cosine scheduling to ensure stable convergence. To enhance computational performance, we integrated DeepSpeed Stage 2 for memory optimization and Flash Attention 2 for efficient attention computation, thereby significantly improving training throughput while maintaining model accuracy. Finally, after 75 hours of training on 8 NVIDIA A100 40G GPUs for the PT and SFT phases, the TianHui LLM was successfully constructed.

### Performance comparison of TianHui with other LLMs

To facilitate the comparison of TianHui's performance with other LLMs in TCM application scenarios, 12 domain-specific benchmark test datasets were developed, encompassing a variety of TCM scenarios. including APQ, TCMCD, TCMEE, HFR, APR, HCCA, GCPMI, DHPE, TCMKQA, TCMRC, TLAW, and ADTG (Supplementary Table 1).

To assess the performance of LLMs in answering single-choice questions within the domain of TCM, we constructed the APQ test dataset comprising 2,000 data. We provided a sample data as a template as a system prompt, and provided questions and options for each data as input in each LLM. Notably, we observed significant variations in output formatting capabilities across different LLMs. While most models could be manually calibrated to produce outputs conforming to the required option format (A, B, C, D, or E), a set of LLMs, including Bentsao, BianQue-2, Chatmed, ShenNong, TCMchat, HuatuoGPT2-13B, Lingdan-13B-Base, Lingdan-13B-PR, and Qwen2.5-Math-7B (mainly TCM-related LLMs), demonstrated continuous formatting and content limitations regardless of prompt engineering attempts. The results indicated that Deepseek-R1 and TianHui outperformed the other LLMs, with accuracy rates of 0.836 and 0.811, respectively, while the accuracy rates of all other LLMs were below 0.8 (Figure 3A). TCMCD held significant clinical and theoretical value, and evaluating a model's ability to classify syndrome patterns (treatment based on syndrome differentiation) provided a robust measure of its diagnostic reasoning performance. The results demonstrated that Deepseek-R1, Qwen2.5-72B-Instruct, and TianHui outperformed the other LLMs, with accuracy rates of 0.838, 0.782, and 0.754, respectively, while the accuracy rates of all other LLMs were below 70% (Figure 3B). Similarly, we still found that regardless of modifying the prompts, some LLMs, including Bentsao, BianCang-Qwen2.5-7B-Instruct, BianQue-2, Chatmed, Lingdan-13B-Base, Lingdan-13B-PR, Qwen2.5-Math-7B, and ShenNong (mainly TMC-related LLMs), could not output the format and content correctly. To evaluate the performance of LLM on the TCM-related entity extraction task, we constructed the TCMEE test dataset containing 480 data. The results for the TCMEE test dataset showed that TianHui (Precision: 0.825, Recall: 0.770, F1-score: 0.779), TCMchat (Precision: 0.659, Recall: 0.799, F1-score: 0.697), and Qwen2.5-32B-Instruct (Precision: 0.735, Recall: 0.677, F1-score: 0.691) significantly outperformed the other LLMs (Figure 3C). We found that the performance of TianHui was best in the entity recognition task.

Figure 3 Performance comparison of TianHui with other LLMs in APQ, TCMCD, and TCMEE scenarios.

(A and B) The performance of LLM was evaluated on the APQ and TCMCD datasets with accuracy. (C) The performance of LLM was evaluated on the TCMEE dataset with precision, recall, and F1 score.

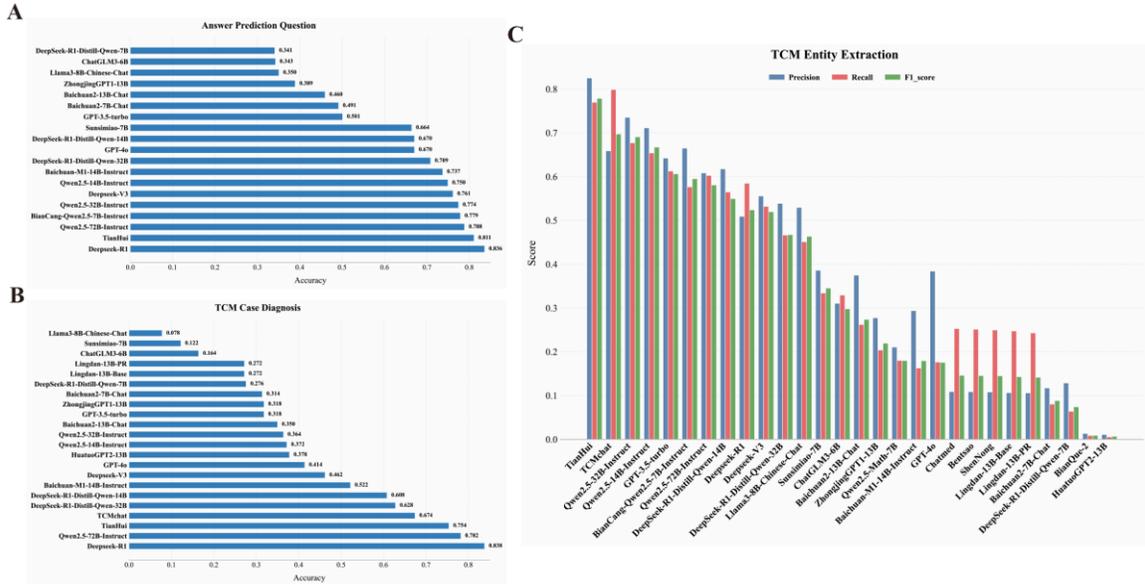

Intelligent recommendation of drugs and acupoints is important in TCM, and we constructed the HFR test set containing 500 data and the APR test set containing 350 data. The results for the HFR test set showed that TCMchat (MRR: 0.326, Precision@3: 0.275, Recall@3: 0.314, HR@3: 0.476, nDCG: 0.283), TianHui (MRR: 0.311, Precision@3: 0.249, Recall@3: 0.204, HR@3: 0.362, nDCG: 0.182), and Deepseek-V3 (MRR: 0.180, Precision@3: 0.135, Recall@3: 0.021, HR@3: 0.064, nDCG: 0.037) significantly outperformed the other LLMs (Figure 4A). The results for the APR test set showed that TianHui (MRR: 0.895, Precision@3: 0.766, Recall@3: 0.267, HR@3: 0.880, nDCG: 0.640), Qwen2.5-32B-Instruct (MRR: 0.867, Precision@3: 0.636, Recall@3: 0.224, HR@3: 0.711, nDCG: 0.527), and GPT-4o (MRR: 0.858, Precision@3: 0.676,

Recall@3: 0.238, HR@3: 0.817, nDCG: 0.607) significantly outperformed the other LLMs (Figure 4B).

Figure 4 Performance comparison of TianHui with other LLMs in HFR, and APR scenarios.

(A and B) In the HFR and APR datasets, the performance of LLMs was evaluated by MRR, Precision@3, Recall@3, HR@3, and nDCG.

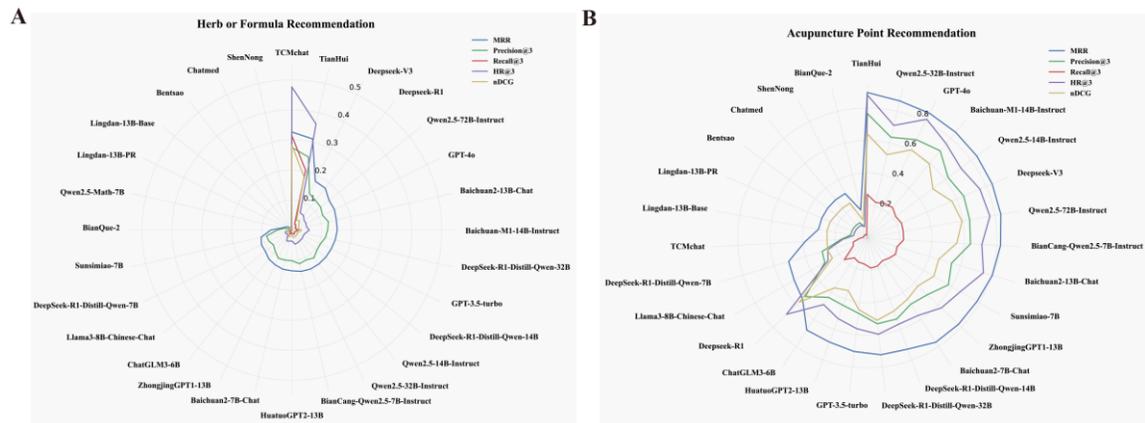

We further employed the natural language generation capacity of LLMs to evaluate their comprehension of TCM knowledge across seven benchmark datasets: HCCA, TLAW, GCPMI, DHPE, TCMKQA, TCMRC, and ADTG. The results for the HCCA test set showed that Deepseek-V3 (BLEU-1: 0.391, BLEU-4: 0.181, BERTScore: 0.765, ROUGE-1: 0.428, ROUGE-2: 0.177, ROUGE-L: 0.325, METEOR: 0.279), and TianHui (BLEU-1: 0.432, BLEU-4: 0.184, BERTScore: 0.755, ROUGE-1: 0.404, ROUGE-2: 0.153, ROUGE-L: 0.300, METEOR: 0.283) significantly outperformed the other LLMs (Figure 5A). The results for the TLAW test set showed that Qwen2.5-32B-Instruct (BLEU-1: 0.357, BLEU-4: 0.200, BERTScore: 0.727, ROUGE-1: 0.341, ROUGE-2: 0.122, ROUGE-L: 0.222, METEOR: 0.215), Deepseek-V3 (BLEU-1: 0.418, BLEU-4: 0.246, BERTScore: 0.716, ROUGE-1: 0.316, ROUGE-2: 0.102, ROUGE-L: 0.204,

METEOR: 0.204), and TianHui (BLEU-1: 0.466, BLEU-4: 0.272, BERTScore: 0.716, ROUGE-1: 0.328, ROUGE-2: 0.126, ROUGE-L: 0.225, METEOR: 0.240) significantly outperformed the other LLMs (Figure 5B). The results for the the GCPMI test set showed that TianHui outperformed other models with BLEU-1 (0.603), BLEU-4 (0.476), BERTScore (0.844), ROUGE-1 (0.633), ROUGE-2 (0.515), ROUGE-L (0.599) and METEOR (0.598) (Figure 5C). The results for the DHPE test set showed that Deepseek-V3 (BLEU-1: 0.435, BLEU-4: 0.169, BERTScore: 0.749, ROUGE-1: 0.369, ROUGE-2: 0.134, ROUGE-L: 0.279, METEOR: 0.272), and TianHui (BLEU-1: 0.324, BLEU-4: 0.135, BERTScore: 0.739, ROUGE-1: 0.350, ROUGE-2: 0.135, ROUGE-L: 0.275, METEOR: 0.227) significantly outperformed the other LLMs (Figure 5D). The results for the the TCMKQA test set showed that TianHui outperformed other models with BLEU-1 (0.432), BLEU-4 (0.198), BERTScore (0.779), ROUGE-1 (0.447), ROUGE-2 (0.194), ROUGE-L (0.353) and METEOR (0.309) (Figure 5E). The results for the the TCMRC test set showed that TianHui outperformed other models with BLEU-1 (0.659), BLEU-4 (0.624), BERTScore (0.893), ROUGE-1 (0.774), ROUGE-2 (0.716), ROUGE-L (0.760) and METEOR (0.750) (Figure 5F). The results for the the ADTG test set showed that TianHui outperformed other models with BLEU-1 (0.659), BLEU-4 (0.492), BERTScore (0.812), ROUGE-1 (0.561), ROUGE-2 (0.353), ROUGE-L (0.503) and METEOR (0.566) (Figure 5G).

Figure 5 Performance comparison of TianHui with other LLMs in TCMKQA, GCPMI, TCMRC, DHPE, HCCA, ADTG, and TLAW scenarios.

(A-G) In the TCMKQA, GCPMI, TCMRC, DHPE, HCCA, ADTG, and TLAW datasets, the performance of LLMs was evaluated by BLEU-1, BLEU-4, BERTScore, ROUGE-1, ROUGE-2, ROUGE-L, and METEOR.

In summary, the current benchmark results showed that TianHui had the best overall performance across 12 different scenarios, with it in the top three for each metric in six scenarios (APQ, TCMCD, HFR, HCCA, DHPE, and TLAW test datasets) and first for each metric in another six scenarios (TCMEE, APR, GCPMI, TCMKQA, TCMRC, and ADTG test datasets). Remarkably, although some of the TCM-related LLMs performed well on some tasks, the overall performance of the overall TCM-related LLMs was relatively poor. For instance, while TCMchat exhibited superior performance on the TCMEE, HFR, and TCMRC datasets, this observed advantage potentially originated from data leakage, as these evaluation sets were presumably incorporated in the model's training corpus. As another example, no matter how the prompts were modified, a large number of TCM-related LLMs, including Bentsao, BianCang-Qwen2.5-7B-Instruct, BianQue-2, Chatmed, Lingdan-13B-Base, Lingdan-13B-PR, and ShenNong, were unable to correctly output format and content in APQ, and TCMCD datasets. Moreover, we also find that the larger the number of parameters, the later the release, and the more Chinese characters contained in the corpus, the better the performance of the LLMs on these test sets.

**Ablation result**

We performed extensive ablation of TianHui and summarised our findings below (Table 3). For the hyper-parameters LoRA rank and LoRA alpha, we empirically guaranteed a ratio of LoRA alpha/LoRA rank of 2 for the ablation experiments. The best model performance was achieved when the LoRA rank was 128 and the LoRA alpha was 256, with a positive correlation with model performance. For the hyper-parameter epoch, we selected 2, 4, and 6 for our experiments, and the results showed that the best performance of the model was achieved when epoch was 4. For the hyper-parameter dropout, we selected 0, 0.2, and 0.4 for our experiments, and the results indicated that the best performance of the model was achieved with a dropout of 0.2. For the hyper-parameter max length, we chose 256, 512, 1025 and 2024 in our experiments, and the results demonstrated that the best performance of the model was achieved with a max length of 2024. Finally, the hyper-parameters of TianHui were determined to be LoRA rank of 128, LoRA alpha of 256, epoch of 4, dropout of 0.2, and max length of 2048.

Table 3: Ablation study of TianHui, evaluating the impact of LoRA configurations (rank, alpha), training epochs, dropout rates, and sequence length on performance across 12 TCM-related tasks (APQ, TCMCD, TCMEE, HFR, APR, HCCA, GCPMI, DHPE, TCMKQA, TCMRC, TLAW, and ADTG). Results demonstrate that optimal settings (LoRA rank=128, alpha=256, epoch=4, dropout=0.2, max length=2048) achieve the highest accuracy and Accuracy/F1/MRR/BLEU scores.

| Ablation | APQ(Accuracy) | TCMCD(Accuracy) | TCMEE(F1_score) | HFR(MRR) | APR(MRR) | HCCA(BLEU-1) | TLAW(BLEU-1) | GCPMI(BLEU-1) | DHPE(BLEU-1) | TCMKQA(BLEU-1) | TCMRC(BLEU-1) | ADTG(BLEU-1) |
|---|---|---|---|---|---|---|---|---|---|---|---|---|
| TianHui(LoRA rank=128 LoRA alpha=256 Epoch=4 Dropout=0.2 | **0.811** | **0.754** | **0.779** | **0.311** | **0.895** | **0.432** | **0.466** | **0.603** | **0.324** | **0.432** | <u>0.659</u> | **0.659** |
| Max length=2048) | | | | | | | | | | | | |
| LoRA rank=8 LoRA alpha=16 | 0.642 | 0.490 | 0.587 | 0.189 | 0.742 | 0.394 | 0.256 | 0.412 | 0.219 | 0.358 | 0.613 | 0.623 |
| LoRA rank=16 LoRA alpha=32 | 0.673 | 0.396 | 0.675 | 0.120 | 0.733 | 0.393 | 0.337 | 0.289 | 0.237 | 0.349 | 0.645 | 0.634 |
| LoRA rank=32 LoRA alpha=64 | 0.722 | 0.598 | 0.669 | 0.199 | 0.725 | 0.402 | 0.260 | 0.517 | 0.247 | 0.385 | 0.643 | 0.626 |
| LoRA rank=64 LoRA alpha=128 | 0.759 | 0.548 | 0.724 | 0.207 | 0.854 | 0.400 | 0.326 | 0.555 | 0.268 | 0.397 | 0.639 | 0.648 |
| Epoch=2 | 0.750 | 0.744 | 0.772 | 0.250 | 0.876 | 0.395 | 0.416 | 0.562 | 0.282 | 0.380 | 0.641 | 0.636 |
| Epoch=6 | 0.811 | 0.686 | 0.774 | 0.267 | 0.861 | 0.407 | 0.386 | 0.537 | 0.245 | 0.323 | 0.652 | 0.494 |
| Dropout=0 | 0.803 | 0.612 | 0.767 | 0.249 | 0.839 | 0.406 | 0.403 | 0.543 | 0.257 | 0.312 | **0.660** | 0.547 |
| Dropout=0.4 | 0.798 | 0.682 | 0.743 | 0.255 | 0.855 | 0.408 | 0.400 | 0.529 | 0.251 | 0.415 | 0.650 | 0.568 |
| Max length=256 | 0.726 | 0.368 | 0.601 | 0.202 | 0.810 | 0.400 | 0.343 | 0.396 | 0.279 | 0.342 | 0.557 | 0.629 |
| Max length=512 | 0.747 | 0.602 | 0.685 | 0.200 | 0.780 | 0.405 | 0.404 | 0.458 | 0.258 | 0.346 | 0.649 | 0.616 |
| Max length=1024 | 0.782 | 0.748 | 0.734 | 0.229 | 0.894 | 0.404 | 0.355 | 0.555 | 0.252 | 0.390 | 0.651 | 0.577 |

## Discussion

Despite the accumulation of extensive generalized knowledge by LLMs, greater precision was required regarding the terminology, context, and knowledge structure of a particular industry or domain[31]. Through fine-tuning on a dataset specific to a particular vertical, the LLM was enabled to better understand and generate content specialized in that domain, thereby enhancing its accuracy and relevance within that domain[32]. In this study, we trained an LLM using a combination of PT and SFT phases and successfully developed an LLM in the field of TCM called TianHui. TianHui significantly outperformed known LLMs in 12 different TCM-related scenarios, including APQ, TCMCD, TCMEE, HFR, APR, HCCA, GCPMI, DHPE, TCMKQA, TCMRC, TLAW, and ADTG. Based on the current findings, TianHui, which is based on the fine-tuning of generic large-scale models, performed well in various application tasks of TCM and has great potential.

The evaluation of LLM was crucial, and the accuracy of its results directly determined the credibility of the performance of the LLM[33]. However, there were significant limitations in the current assessment of LLMs for the TCM domain [34]. For example, TCM-GPT was evaluated on two datasets: TCM-EXAM, which focused on standardized exam questions to assess theoretical knowledge; and TCM-EHR, which was an electronic health record collection covering multiple clinical cases[35]. TCM-FTP only used DigestDS, a specialized dataset of digestive system disease prescriptions extracted from real clinical records for testing symptom to prescription predictions [35]. Hengqin-RA-v1 was limited to HQ-GCM-RA-C1, a dataset specifically designed for rheumatoid arthritis research that included clinical data and treatment plans for the condition[37]. The five tasks employed for evaluation in TCMChat, which included multiple-choice questions, reading comprehension, entity extraction, medical case diagnosis, and herbal or prescription recommendations,,ignored key areas such as pharmacological reasoning, chemical analysis, and scientific text generation[37]. Currently, the evaluation datasets of LLM in TCM mainly focused on the two major scenarios of clinical practice and medical education, but generally ignored the diverse needs of real-world application scenarios and the professional data support required for research and innovation[40]. Our evaluation framework addressed this gap by integrating 12 comprehensive evaluation datasets that covered not only a full range of downstream tasks, but also three key areas: clinical practice, medical education, and scientific research, making LLM a versatile tool for promoting innovation in TCM in education, academic research, and healthcare.

In comparative and ablation experiments, the TianHui demonstrated its high efficiency and accuracy in handling TCM-related tasks, thanks to its outstanding performance on 12 evaluation datasets and multiple key hyper-parameters. However, some TCM-specialized LLMs had significant shortcomings in the accuracy of output format and content, which prompted us to deeply consider the factors affecting model performance. In the study of the Biancang, the author trained models with two parameter sizes, 7B and 14B (based on Qwen2/2.5). The author compared the models with different parameter sizes and found that in the disease diagnosis task (TCMDD-BC), the accuracy of the 14B model (86.33%) was significantly higher than that of the 7B model (87.87%). In the MLEC-TCM exam task, the zero-shot accuracy of the 14B model (92.29%) was also better than that of the 7B model (90.22%) [40]. Therefore, combined with our experimental results, we speculated that the reason for the inability of LLMs to be generated according to instructions might be related to the small parameter size of LLMs. These results consistently demonstrated that larger parameter sizes lead to better performance in TCM-specific tasks. The increased parameter count enhanced the LLM's capacity to capture intricate TCM knowledge patterns and clinical reasoning pathways, which was particularly crucial for handling the complexity and nuance inherent in TCM applications.

Our ablation experiment results indicated that the settings of LoRA rank and LoRA alpha had a significant impact on model performance. A higher LoRA rank of 128 and a larger LoRA alpha of 256 could bring better overall performance, such as achieving 0.811, 0.779, and 0.895 in APQ, TCMEE, and APR indicators, respectively, significantly better than configurations with low rank of 8 and low alpha of 16, which were only 0.642, 0.587, and 0.742 in these indicators. This indicated that a larger LoRA parameter scale could effectively enhance the model's ability to capture task features, which was a key advantage in improving model performance. However, this advantage came at a cost of higher GPU memory overhead; for example, configurations with rank=128 and alpha=256 required more computing resources compared to rank=8 and alpha=16. In addition, as the training epochs increased (from 2 to 4), the model performance improved on most metrics (such as APQ increasing from 0.750 to 0.811), but further increasing to 6 epochs caused some metrics (such as TLAW and DHPE) to show a decrease, which might be related to overfitting. The adjustment of dropout rate (from 0.2 to 0.4) had a relatively small impact on performance, but completely removing dropout (dropout = 0) led to a decrease in metrics such as TCMCD, verifying the necessity of regularization. The maximum length setting experiment showed that a longer context window (such as 2048) could significantly improve the performance of tasks such as TCMCD (0.754 vs. 256 length 0.368) and APR (0.895 vs. 0.810), indicating that complex tasks required more comprehensive contextual information. Overall, increasing LoRA rank and LoRA alpha could significantly improve model performance, but it required more GPU memory consumption.

While TianHui performed well in a wide range of scenarios, there were still issues that needed to be addressed. On the one hand, TianHui lacked multi-modal processing capabilities, creating a gap with real-world clinical practice. The four diagnostic methods in TCM (observation, auscultation-olfaction, inquiry, and palpation) required multi-modal data integration (visual, auditory, textual, and tactile), which aligned naturally with multi-modal model architectures. Therefore, our future research will focus on building a multi-modal data that integrates heterogeneous data such as clinical information, image data, voice recordings, textual medical records, and biosignals, in order to train MLLMs, which will expand the cognitive dimensions of the models and enhance their application value in TCM clinical research. On the other hand, due to the limitation of insufficient computational resources, the training of LLMs with 14B parameters was only completed so far. However, it should be noted that the number of model parameters was one of the key factors affecting the performance of LLMs. According to the current research progress, the number of parameters in mainstream LLMs generally reached a much larger scale (e.g., tens of billions or even hundreds of billions of levels). In subsequent research, we will further integrate computational resources, optimize resource allocation, and expand the training data size to carry out LLM training with larger parameter scales, with the aim of obtaining better model performance.

## Conclusion

In conclusion, we successfully developed TianHui, a professional privatised LLM for TCM domain. The benchmark test data showed that TianHui demonstrated excellent performance in 12 TCM-related application scenarios. It ranked in the top three in each evaluation index in six test datasets: APQ, TCMCD, HFR, HCCA, DHPE, and TLAW. Meanwhile, it achieved optimal performance in all indicators of the six test data sets: TCMEE, APR, GCPMI, TCMKQA, TCMRC, and ADTG. Through continuous technology iteration and accumulation of TCM featured data, TianHui not only significantly improved the accuracy and professionalism of TCM knowledge process, but also provided an intelligent solution for the systematic inheritance and large-scale application of TCM knowledge.

## Acknowledgements

We are very grateful to Sichuan Huixin Intelligent Computing Technology Co., Ltd. for providing the equipment and technical support (8 NVIDIA A100 40G GPUs) for this research.

## Data Availability

All data generated or analyzed during this study are included in this published article. Our code, data, and models are all open-sourced on GitHub (https://github.com/JYfantast/TianHui) and HuggingFace (https://huggingface.co/JYfantast/TianHui).

## Conflicts of Interest

None declared.

## Declaration of Generative AI and AI-assisted technologies in the writing process

During the preparation of this work, the author(s) used the AI-based language model service provided by OpenAI (ChatGPT) to enhance the language and readability of the manuscript. After utilizing this tool, the author(s) thoroughly reviewed and edited the content as necessary and take(s) full responsibility for the content of the publication.

## Abbreviations

ADTG: Abstract-Driven Topic Generation

AI: Artificial Intelligence

APQ: Answer Prediction Question

APR: Acupuncture Point Recommendation

BLEU: Bilingual Evaluation Understudy

CNKI: China National Knowledge Infrastructure

CPM: Chinese Patent Medicine

CPMI: Chinese Patent Medicine Instructions

DHPE: Description of Herbal Pharmacological Effect

EPUB: Electronic Publication

FFT: Full Fine-Tuning

GCPMI: Generation of Chinese Patent Medicine Instruction

HCCA: Herbal Chemical Composition Analysis

HFR: Herb or Formula Recommendation

LLM: Large Language Model

LR: Learning Rate

METEOR: Metric for Evaluation of Translation with Explicit Ordering

MLLM: Multimodal Large Language Model

MRR: Mean Reciprocal Rank

nDCG: Normalized Discounted Cumulative Gain

NLP: Natural Language Processing

OCR: Optical Character Recognition

PDF: Portable Document Format

PEFT: Parameter-Efficient Fine-Tuning

PLM: Pre-trained Language Model

PT: Pre-Training

Q&A: Question and Answer

QLoRA: Quantized Low-Rank Adaptation

RL: Reinforcement Learning

RLHF: Reinforcement Learning from Human Feedback

ROUGE: Recall-Oriented Understudy for Gisting Evaluation

SFT: Supervised Fine-Tuning

TCM: Traditional Chinese Medicine

TCMCD: TCM Case Diagnosis

TCMEE: TCM Entity Extraction

TCMKQA: TCM Knowledge Questions and Answer

TCMRC: TCM Reading Comprehension

TLAW: Topic-Led Abstract Writing